\documentclass[lettersize,journal]{IEEEtran}
\usepackage{amsmath,amsfonts}
\usepackage{array}
\usepackage[caption=false,font=normalsize,labelfont=sf,textfont=sf]{subfig}
\usepackage{textcomp}
\usepackage{stfloats}
\usepackage{url}
\usepackage{verbatim}
\usepackage{graphicx}
\usepackage{cite}
\usepackage{colortbl}
\usepackage[table]{xcolor}
\usepackage{xcolor}%
\usepackage{algorithm}
\usepackage{algpseudocode}%
\usepackage{multirow}%
\usepackage{bm}

\usepackage{booktabs}%
\def\BibTeX{{\rm B\kern-.05em{\sc i\kern-.025em b}\kern-.08em
    T\kern-.1667em\lower.7ex\hbox{E}\kern-.125emX}}
\usepackage{balance}
\begin{document}
\title{Functional Subnetwork Localization for Efficient Continual Learning in Large Language Models}
\author{Shuoran Jiang,
Wentao Zhu,
Yang Xiang,
Dongfeng Li,
Yunan Zhang,
Mengchen Zhao,
Yuhui Guo,
Kai Chen,
Youcheng Pan,
Xiangping Wu,
Qingcai Chen
% First affiliation
\thanks{
Shuoran Jiang, Wentao Zhu, Dongfeng Li, Yunan Zhang, Mengchen Zhao, Yuhui Guo, Xiangping Wu and Qingcai Chen are with the School of Computer Science and Technology,
Harbin Institute of Technology (Shenzhen),
Shenzhen, China
(e-mail: jiangshuoran@hit.edu.cn;
zhangyunan@stu.hit.edu.cn;
zhaozhao436@gmail.com;
25B951070@stu.hit.edu.cn;
wuxiangping@stu.hit.edu.cn;
qingcai.chen@hit.edu.cn).
}%

\thanks{
Dongfeng Li is with the School of Science,
Harbin Institute of Technology (Shenzhen),
Shenzhen, China
(e-mail: lidongfeng@hit.edu.cn).
}%

% Second affiliation
\thanks{
Wentao Zhu is with the
China Academy of Engineering Physics,
Mianyang, China
(e-mail: zhuwentao66666@gmail.com).
}%

% Third affiliation
\thanks{
Yang Xiang, Youcheng Pan and Qingcai Chen are with the
Peng Cheng Laboratory,
Shenzhen, China
(e-mail: xiangy@pcl.ac.cn;
panyoucheng4@gmail.com;
qingcai.chen@hit.edu.cn).
}%

% Fourth affiliation
\thanks{
Kai Chen is with the
Huaqiao University,
Xiamen, China
(e-mail: chenkai.cs@hqu.com).
}%

% Corresponding author
\thanks{
Corresponding author: Qingcai Chen and Xiangping Wu
(e-mail: qingcai.chen@hit.edu.cn;
wuxiangping@stu.hit.edu.cn).
}}

\markboth{Journal of \LaTeX\ Class Files,~Vol.~18, No.~9, September~2020}%
{Endogenous cognitive modularity in large language models enables human-like continual adaptation}

\maketitle

\begin{abstract}
Biological intelligence continually adapts to new information while retaining prior skills through localized plasticity.
In contrast, continual learning in large language models (LLMs) remains vulnerable to catastrophic forgetting.
A promising direction is to identify functionally specialized subnetworks within LLMs that enable targeted and resource-efficient adaptation.
Using gradient-attribution profiling, we localize cognition-associated functional subnetworks exhibiting extremely low mutual overlap.
Building on this finding, we propose Function-isolated Memory-Efficient Optimization (CiMEO), a continual learning framework that routes each training example to its corresponding functional subnetwork, restricting parameter updates and optimizer states to selected regions.
Across 1.4 million sequential instructions, CiMEO achieves the highest average performance among the compared methods on five generative benchmarks spanning programming, examination-style tasks, and mathematical reasoning, while mitigating catastrophic forgetting.
Subspace-overlap and principal-angle analyses further indicate that localized updates help preserve pre-trained representations, highlighting functional subnetwork localization as a practical approach to efficient and stable continual learning in LLMs.
\end{abstract}


\section{Introduction}
\IEEEPARstart{L}{arge} language models (LLMs) are increasingly deployed as long-lived knowledge compilation and content generation systems \cite{naveed2025comprehensive}, which must be updated after pre-training to incorporate new knowledge, align with evolving preferences and expand to new domains \cite{shi2024continual, veniat2021efficient}.
This is the research content of continual learning for LLMs, the central challenge in which is not merely learning new data, but preserving previously acquired knowledge \cite{wang2024comprehensive}.
However, new updates in LLMs' continual learning usually overwrite earlier skills and degrade general, which induces the catastrophic forgetting \cite{mccloskey1989catastrophic,french1999catastrophic}.
This forgetting problem reflects a fundamental tension that LLMs must remain plastic enough to acquire new behaviors, while stay stable enough to retain old ones \cite{lai2025pareto}.
For long-term deployed LLMs, maintaining stability is particularly crucial because it ensures the coherence of user experience, and it is also particularly challenging because updates are typically applied to global parameters, allowing interference to accumulate throughout the entire deployment lifecycle.
Practical constraints further exacerbate the problem, as standard optimizers maintain dense states. Therefore, attempts to improve stability through repeated trials can become prohibitively expensive in terms of memory and computation \cite{lv2024full}.

\begin{figure*}[htbp]
	\centering
	\includegraphics[width=1.0\textwidth]{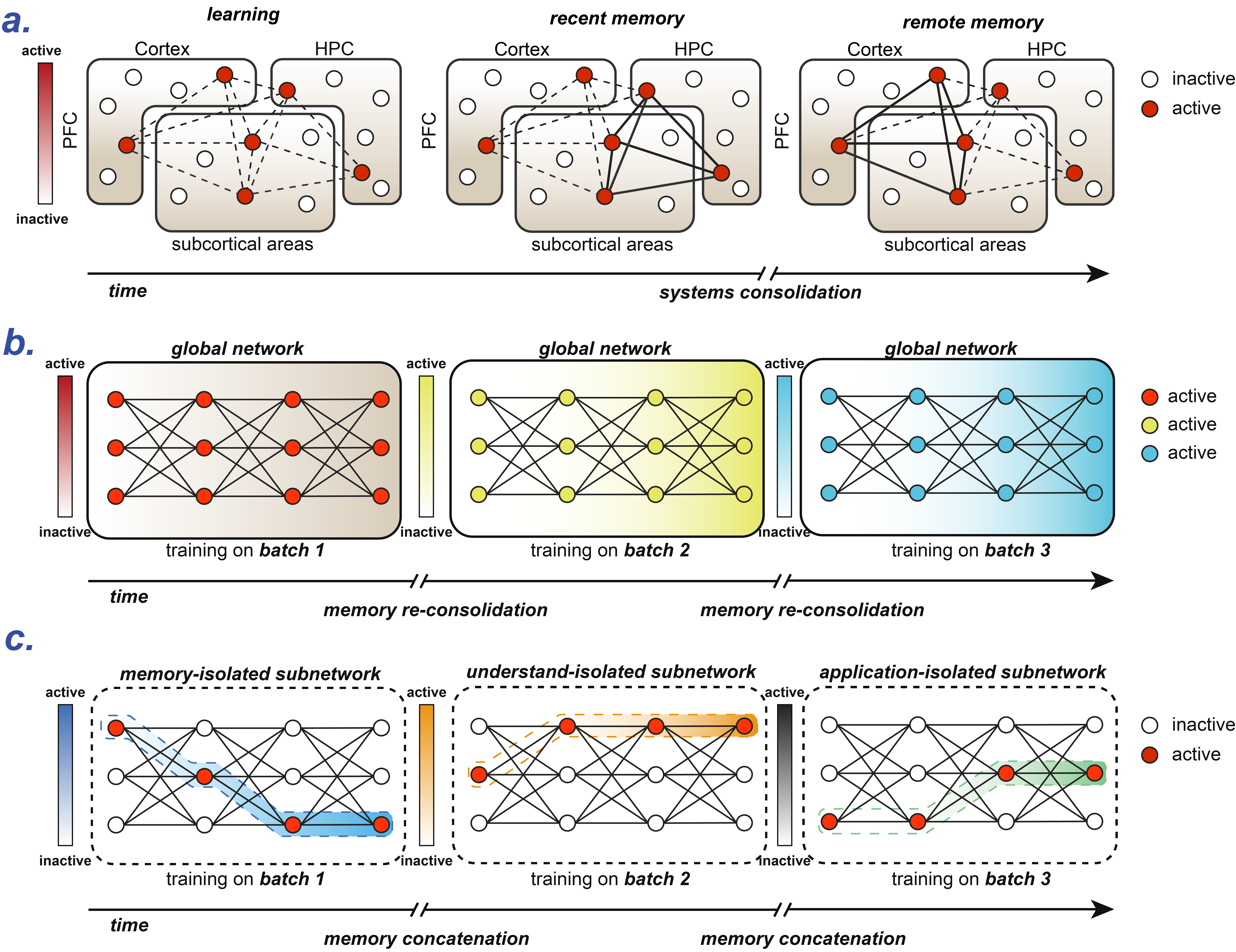}
	\caption{localized plasticity in human barin vs. global plasticity in LLMs. (\textbf{a}), Learning recruits and co-activates sparse neuronal ensembles (circles; color indicates activity) across memory-relevant regions, including the hippocampus (HPC), cortex, and subcortical areas, forming an engram. During the first hours to days after learning, synaptic connectivity is preferentially strengthened among HPC engram cells and between HPC and other regions, enabling recent memory expression. Over subsequent days to weeks, systems consolidation supported by hippocampo–cortical replay and synchronized activity (notably involving prefrontal cortex, PFC) promotes maturation of cortical engram cells. This process drives strengthening of synaptic connectivity between cortical-cortical and cortical-subcortical engram cells and a progressive shift in the engram cell network toward a higher dependency on cortical engram cells for remote memory expression. (\textbf{b}), Global parameter updates across sequential batches bring memory re-consolidation. During training on successive batches, learning is typically implemented via dense, global weight updates across the network. Lacking a brain-like mechanism of sparse engram allocation and multi-stage consolidation, new batches are absorbed by rewriting shared parameters, which can interfere with or overwrite earlier representations. As a result, the model’s ``memory'' is continually reconfigured by the order of batches, reflecting a stability–plasticity trade-off and potential forgetting under sequential learning.}
	\label{fig:brain_w_llm}
\end{figure*}

Existing continual learning methods improve stability mainly by (\romannumeral1) regularizing parameter drift, (\romannumeral2) rehearsing stored past examples, or (\romannumeral3) distilling behavior from an earlier model \cite{grossberg2013adaptive}.
However, these strategies become difficult to scale for LLMs' continual learning.
Regularization methods, such as elastic weight consolidation (EWC) \cite{kirkpatrick2016overcoming,zhang2024elastic}, require estimating and storing parameter importance at task boundaries, which increases extra computation and state beyond standard training.
Rehearsal-based approaches mitigate forgetting by replaying data from a memory buffer \cite{liu2025contextual}, but this introduces a persistent memory footprint.
Their effectiveness also degrades when the buffer is small, and will be amplified in large-scale parameter settings.
Distillation-based methods typically require keeping a previous model and performing additional forward passes to match prior predictions \cite{li2024continual}, further increasing compute overhead.
Crucially, many of these stability-oriented methods still assume global parameter updates with Adam-like optimizers, whose first- and second-moment accumulators add substantial state memory, making repeated post-training prohibitively expensive under strict budgets \cite{wang2025parameter,chen2025memory}.
As surveyed in recent reviews \cite{tang2025training,wan2024efficient}, this cost–stability mismatch remains a key obstacle for LLMs' long-horizon deployment \cite{gupta2025effective}.

Biological learning offers a useful contrast: continual adaptation in the human brain is thought to local synaptic plasticity \cite{josselyn2020memory,yuste2024neuronal}, where learning is expressed through the selective recruitment of task-relevant neuronal ensembles while other circuits remain comparatively stable \cite{josselyn2020memory,yuste2024neuronal,josselyn2018memory,lavi2023local,larsen2015synapse} as shown in Fig.\ref{fig:brain_w_llm}.
This suggests a concrete design principle for the continual learning of Large Language Models (LLMs), which involves identifying reusable subnetworks according to a certain principle.
Every incoming sample is allocated to a specific subnetwork to update parameters in those isolated subspaces, so that new learning is localized rather than globally disruptive.

Motivated by this principle, we investigate whether pretrained
LLMs exhibit functionally specialized parameter regions that
can support localized adaptation during continual learning.
Guided by Bloom's taxonomy \cite{krathwohl2002revision},
we employ gradient-attribution profiling to estimate parameter
importance across six cognitive categories and localize the
corresponding functional subnetworks.
The identified subnetworks exhibit remarkably low mutual overlap,
averaging below 0.002\% across cognitive categories.
This observation motivates a functional subnetwork localization
strategy that restricts parameter updates to cognition-relevant
regions while minimizing interference with previously acquired
knowledge.

Building on this observation, we propose
\textbf{C}ognition-\textbf{i}solated
\textbf{M}emory-\textbf{E}fficient
\textbf{O}ptimization (CiMEO), a framework for efficient
continual learning in LLMs.
CiMEO routes each incoming instruction to its corresponding
functional subnetwork, selectively updates relevant parameters,
and maintains optimizer states only for the selected indices.
By integrating cognition-guided routing, functional subnetwork
localization, and memory-efficient optimization, CiMEO aims to
mitigate cross-skill interference and catastrophic forgetting
while reducing the optimizer-state overhead of continual
adaptation without relying on replay buffers.

Overall, the contributions of this study are:
\begin{itemize}
    \item \textbf{Functional subnetwork localization:}
    We identify cognition-associated functional subnetworks
    in pretrained LLMs through gradient-based attribution,
    revealing an average overlap below 0.002\% across
    cognitive categories and providing empirical evidence
    of localized parameter importance.

    \item \textbf{Cognition-guided continual learning:}
    We propose CiMEO, a continual learning framework that
    routes incoming instructions to functionally relevant
    subnetworks and restricts parameter updates to selected
    regions to mitigate interference between successive
    learning stages.

    \item \textbf{Memory-efficient optimization and analysis:}
    We integrate sparse, state-serialized optimization
    to reduce optimizer-state requirements and investigate
    the stability of pretrained representations through
    subspace-overlap and principal-angle analyses.
\end{itemize}

\section{Related Works}
\label{sec:related_works}

\subsection{Continual Learning and Interference Mitigation}
Continual learning seeks to preserve previously acquired capabilities while incorporating information from a non-stationary sequence of tasks. Early approaches address this stability--plasticity dilemma through complementary mechanisms. Elastic weight consolidation (EWC) penalizes changes to parameters estimated to be important for earlier tasks \cite{kirkpatrick2016overcoming}, whereas Learning without Forgetting preserves previous predictions through distillation on newly available data \cite{li2017learning}. Rehearsal-based methods revisit earlier examples; for language models, LAMOL instead generates pseudo-examples of previous tasks during subsequent training \cite{sun2020lamal}. Such strategies can reduce forgetting, but their regularization statistics, replay generation, or teacher-based constraints introduce costs or dependencies that differ from direct parameter isolation.

Subspace-constrained adaptation offers another route to mitigating interference. O-LoRA assigns successive language-model tasks to approximately orthogonal low-rank subspaces without storing rehearsal data \cite{wang2023orthogonal}. InfLoRA constructs low-rank update subspaces designed to avoid interference with previous tasks, primarily evaluated in visual continual learning \cite{liang2024inflora}. More recently, CLoRA regularizes LoRA update directions to reduce forgetting during continued LLM training \cite{lu2025controlled}. These approaches motivate limiting the interaction between consecutive updates; in contrast, CiMEO restricts updates to cognition-attributed \emph{weight coordinates} rather than learning or constraining a separate low-rank update for each task.

\subsection{Parameter Isolation, Sparse Adaptation, and Routing}
Parameter-isolation methods allocate distinct portions of model capacity to different learning objectives. PackNet sequentially prunes and retrains a shared network to reserve previously learned weights while freeing parameters for new tasks \cite{mallya2018packnet}. Piggyback learns task-dependent binary masks over fixed pretrained weights \cite{mallya2018piggyback}; Diff Pruning learns sparse task-specific parameter differences in pretrained language models \cite{guo2021parameter}. Rather than modifying backbone weights, Learning to Prompt (L2P) selects and optimizes small prompts from a learned pool, enabling rehearsal-free continual learning in vision settings without task identity at inference time \cite{wang2022learning}. In language-model adaptation, LoRA \cite{hu2022lora} and sparse approaches such as SHiRA \cite{bhardwaj2024sparse} likewise reduce the extent of trainable parameter changes. More directly, PaRSP identifies task-specific functional regions in LLMs and constrains updates using null-space orthogonality to balance retention, adaptation, and remaining capacity \cite{rwChen2026PaRSP}. These results reinforce the value of localized adaptation, while parameter efficiency alone does not imply protection against interference across a sequence of updates.

These studies establish that allocating or selecting a limited set of parameters can preserve useful representations. CiMEO differs in its \emph{allocation granularity and routing criterion}: it groups incoming instructions by six Bloom-inspired cognitive categories \cite{krathwohl2002revision}, identifies attribution-ranked coordinates within the pretrained model, and uses a frozen cognition classifier to route subsequent updates. Thus, the proposed partition is based on cognitive labels rather than only dataset or task identifiers, and it is paired with sparse optimizer-state storage.

\subsection{Functional Localization and Gradient-Based Attribution}
Evidence for localized functionality in Transformers provides a direct motivation for selective updates. Dai \emph{et al.} identified knowledge-associated neurons in pretrained Transformers through attribution and explored targeted modifications of factual knowledge \cite{rwDai2022KnowledgeNeurons}. Wang \emph{et al.} identified task-predictive \emph{skill neurons}, showing that perturbing them can affect task performance \cite{rwWang2022SkillNeurons}. Surgical fine-tuning further demonstrates that restricting adaptation to selected layers can outperform broader updates under certain distribution shifts \cite{rwLee2023Surgical}. These findings support examining \emph{where} learning signals act, but neither neuron attribution nor a sparse mask by itself establishes that a model contains causally independent cognitive modules.

Most closely related to our study, Leng and Xiong \cite{rwLeng2025TaskSpecificNeurons} used gradient attribution to identify task-specific neurons in LLMs, analyzed their overlap, and proposed neuron-level selective fine-tuning for continual learning. Hence, gradient-based identification and localized continual updates should not be regarded as novel in isolation. CiMEO investigates a different organizational axis---a shared taxonomy of cognitive categories---and couples element-wise cognition-specific coordinate selection with instance-level routing and state-serialized AdamW optimization. Our overlap and weight-subspace analyses characterize this particular setting; the observed separation is empirical and should not be interpreted as proof of strict functional independence.

\subsection{Memory-Efficient Optimization for LLM Adaptation}
Reducing interference is only one requirement of sustained LLM adaptation: optimizer moments and gradients can also dominate the training-memory budget. Parameter-efficient techniques such as LoRA \cite{hu2022lora} shrink the trainable state by introducing compact update matrices. LOMO reduces full-parameter fine-tuning memory by fusing gradient computation and parameter updates \cite{lv2024full}, while AdaLOMO further targets memory-efficient optimization \cite{lv2024adalomo}. GaLore instead projects gradients into low-rank spaces to reduce optimizer-state memory while retaining full-parameter learning \cite{zhao2024galore}. GCL further combines group-shared parameter adaptation and gradient-aware regularization to continually fine-tune already-sparse LLMs while preserving their sparsity patterns \cite{rwWang2026GCL}. In comparison, CiMEO focuses on selecting cognition-relevant coordinates within a pretrained dense backbone and storing optimizer states only for those coordinates. Most memory-oriented approaches primarily address the representation or storage of updates, rather than deciding \emph{which cognition-relevant original weights} should change for each incoming instruction.

\begin{figure*}[htbp]
	\centering
	\includegraphics[width=1.\textwidth]{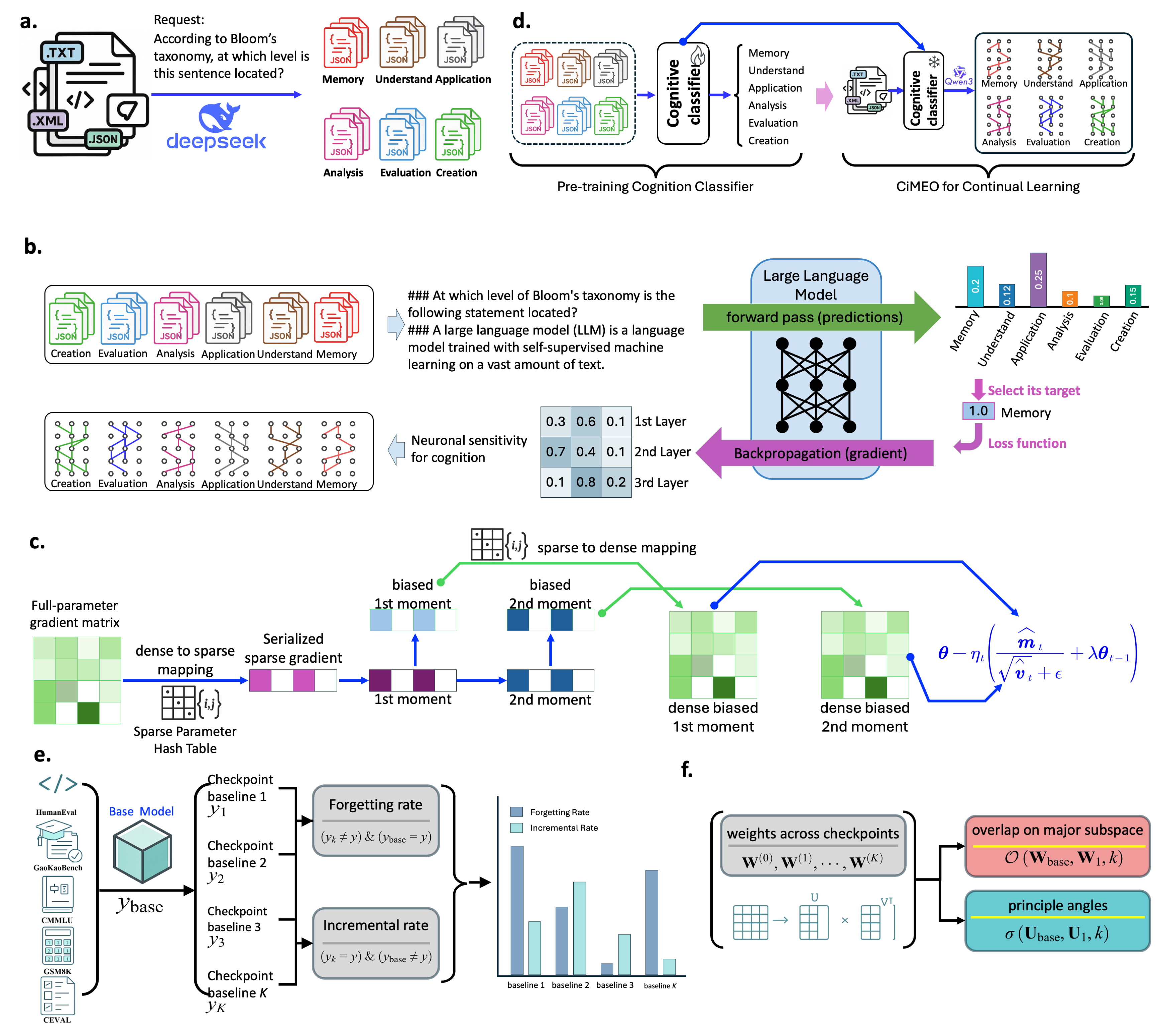}
	\caption{Framework for discovering endogenous cognitive modularity in LLMs and applying it to memory-efficient training. \textbf{a}, We construct a cognitive corpus from the AM-DeepSeek-R1-Distilled-1.4M instruction dataset, where each instruction is annotated with a Bloom’s taxonomy label. \textbf{b}, Cognition-isolated subnetworks are identified via gradient attribution: labelled instructions are fed into the LLM to predict cognition categories, and the gradients accumulated during back-propagation are aggregated to locate cognition-sensitive subnetworks. \textbf{c}, The state-serialized AdamW optimizer maintains a mapping between sparse parameters and their locations in the original dense tensors, so that only cognition-relevant subnetworks are activated and updated during training. \textbf{d}, For continual learning, CiMEO employs a cognitive classifier to assign a cognition label to each input and routes optimization to the corresponding local subnetwork. \textbf{e}, Experiments on five benchmarks report detailed incremental learning and anti-forgetting performance, together with weight-space dynamics during continual learning. \textbf{f}, We further analyze subspace overlap and principal angles across checkpoints to quantify interference and stability in the learned weight subspaces.}
	\label{fig:framework}
\end{figure*}

CiMEO combines both decisions: gradient attribution selects a small set of original weight coordinates, a cognition classifier chooses the relevant set for each example, and S$^2$-AdamW maintains optimizer moments for selected indices. This design aims to control cross-category interference and persistent optimizer-state memory jointly. As with other sparse-update schemes, reductions in stored optimizer states should be distinguished from reductions in activation memory or dense backward-pass computation; the latter depend on the implementation and are not guaranteed by a parameter mask alone.

\section{Method}

This study first collects a cognitive corpus to identify cognition-isolated subnetworks and organizes language generation tasks into Bloom's taxonomy.
Next, to reduce the memory cost of optimization, we propose a state-serialized AdamW.
Finally, we define a continual learning paradigm based on the proposed CiMEO.

\subsection{Cognitive corpus collection}

As shown in Fig.\ref{fig:framework}a,
given a corpus $\mathcal{D}=\{x_i\}_{i=1}^N$ with $N$ plain-text sentences,
we assign each $x_i$ a cognitive label $y_i$ in Bloom's taxonomy,
including \textit{memory}, \textit{understanding}, \textit{application}, \textit{analysis}, \textit{evaluation}, and \textit{creation}.
Let $f(\cdot;\theta)$ denote a frozen instruction-tuned LLM,
each sample $x_i$ is converted into a multiple-choice prompt $Q$,
where the model is asked to select the correct option.

We use beam search to efficiently generate high-probability sequences efficiently.
Beam search maintains multiple candidate sequences (beams) at each decoding step, expands them incrementally and retains only the top-$k$ sequences with the highest normalized scores.
The procedure terminates when all beams generate an end-of-sequence token or reach the maximum length.
For each option $O_j$, its selection frequency $f^{(k)}(x_i; \theta)$ is computed across all runs.
The option with the highest frequency is selected as the final answer:
\begin{equation}
	y_i = \arg \max_{\bm{O}} \lbrace f^{(1)}(x_i; \theta), \cdots, f^{(K)}(x_i; \theta) \rbrace
\end{equation}

Finally, the cognitive Corpus is $\mathcal{D}^{\star} = \lbrace x_i, y_i \rbrace_{i=1}^N$.

\subsection{Cognition-isolated subnetwork identification}
\label{sec:sparsenetfind}

As shown in Fig.~\ref{fig:framework}b, to identify endogenous cognition-isolated subnetworks in LLMs, we employ a gradient attribution approach.
The key idea is to measure the sensitivity of each parameter to a specific cognition category by aggregating the gradient magnitudes induced by cognition-labeled supervision.
The procedure consists of the following four steps.

\textbf{1st Step: Sample selection}
For each cognition label in Bloom's taxonomy, we sample $M$ labeled instances from the cognitive corpus $\mathcal{D}^{\star}$, yielding subsets
$\widehat{\mathcal{D}}_{c}=\{(x_j,y_j)\}_{j=1}^{M}$.

\textbf{2nd Step: Cognition prediction and supervised loss}
Let $f(\cdot;\boldsymbol{\theta})$ denote the frozen LLM,
and each input $x$ is converted into a multiple-choice cognition classification prompt.
Given the gold cognition label $y$, we define the supervision loss as the negative log-likelihood:
\begin{equation}
\mathcal{L}(x,y;\boldsymbol{\theta}) = - \log p_{\boldsymbol{\theta}}(y \mid x).
\end{equation}
where $p_{\boldsymbol{\theta}}(y|x)$ denotes the prediction probability on prompt $x$.

\textbf{3rd Step: Gradient attribution and importance aggregation}
For each parameter $\theta_k \in \boldsymbol{\theta}$, we compute its gradient and use its magnitude as the attribution score.
Specifically, for cognition category c, we aggregate the gradient magnitudes over $\widehat{\mathcal{D}}_c$ to obtain a parameter-importance estimate:
\begin{equation}
A_{\theta_k}^{(c)} = \frac{1}{M}\sum_{m=1}^{M}\left|\frac{\partial}{\partial \theta_k}\mathcal{L}(x_m,y_m;\boldsymbol{\theta})\right|.
\end{equation}

The resulting $A_{\theta_k}^{(c)}$ reflects how strongly parameter $\theta_k$ responds to supervision signals from cognition $c$.

\textbf{4th Step: Subnetwork selection and hash-table storage}
For each weight tensor $\boldsymbol{\theta}$, we rank its elements by $\mathbf{A}^{(c)}$ and select the top p\% as the cognition-isolated subnetwork for category $c$:
\begin{equation}
\boldsymbol{\theta}_{\textit{s}}^{(c)} = \lbrace \theta_k \in \boldsymbol{\theta}\ \big|\ A_{\theta_k}^{(c)} \ge Q_{1-p}(\mathbf{A}^{(c)}) \rbrace,
\end{equation}
where $Q_{1-p}(\mathbf{A}^{(c)})$ denotes the (1-p)-percentile threshold of attribution scores.

To enable efficient sparse updates in the subsequent state-serialized AdamW optimizer, we store the tensor positions of selected parameters in a hash table:
\begin{equation}
\mathcal{H}^{(c)}=\lbrace \operatorname{index}(\theta_k)\ |\ \theta_k \in \boldsymbol{\theta}_{\textit{s}}^{(c)} \rbrace.
\end{equation}

This procedure yields six cognition-specific subnetworks $\{\mathbf{W}_{\textit{s}}^{(c)}\}$, which can then be used for cognition-aware routing and sparse, memory-efficient optimization in continual learning.

\subsection{State-serialized AdamW}

As shown in Fig.\ref{fig:framework}c,
to achieve memory-efficient parameter updating on cognitive isolated subnetworks,
we serialize the parameters and corresponding gradients, moments in AdamW via the hash table $\mathcal{H}^{(c)}$, as illustrated in Fig.\ref{fig:framework}c.

\textbf{Notations} used in the proposed state-serialized AdamW (S$^2$-AdamW) are defined as follows,

\begin{itemize}
    \item $f(\boldsymbol{\theta}) \in \mathbb{R}$ is the loss of LLMs,
    \item $\mathbf{g}_t \in \mathbf{R}^{m\times n}$ is the gradient on weight $\boldsymbol{\theta}$,
    \item $\mathcal{H}^{(c)}$ maps the sparse parameters and positions in cognition $c$ isolated subnetworks,
    \item $\hat{\boldsymbol{g}_t} \xleftarrow{\mathcal{H}^{(c)}} \boldsymbol{g}_t$ serialized sparse gradients,
    \item $\hat{\boldsymbol{m}}_t$ is the exponential moving average (EMA) on $\hat{\boldsymbol{g}}_t$
    \item $\hat{\boldsymbol{s}}_t$ is the EMA on $\hat{\boldsymbol{g}}_t^2$
\end{itemize}

\begin{algorithm}
\caption{\textit{S$^2$-AdamW}: $\boldsymbol{g}^2_t$ indicates the element-wise square $\boldsymbol{g}_t \odot \boldsymbol{g}_t$, $\alpha=0.001$, $\beta_1=0.9$, $\beta_2=0.999$ and $\epsilon=10^{-8}$. All operations are element-wise on vectors, and $\beta_1^t$ and $\beta_2^t$ represent the power $t$.}
\label{alg:auzo}
\begin{algorithmic}[1]
    \State \textbf{Require:} Step size $\alpha$, exponential decay rates $\beta_1,\beta_2 \in \left[ 0,1 \right)$, scattering subnetwork parameter maps $\boldsymbol{\mathcal{H}}$ and objective function $f(\boldsymbol{\theta})$.
    \State \textbf{Initialize:} $\boldsymbol{\theta}_0$, $\boldsymbol{m}_0\gets \mathbf{0}$, $v_0\gets \mathbf{0}$, $t\gets 0$
    \While{$\theta$ not converged}
        \State $t \gets t+1$
        \State $\Bar{\boldsymbol{\theta}}_t \xleftarrow{\boldsymbol{\mathcal{D}}} \boldsymbol{\theta}_t$
        \State $\boldsymbol{g}_t \gets \bigtriangledown _{\boldsymbol{\theta }}f\left( \boldsymbol{\theta }_{t-1} \right) $, \quad $\Bar{\boldsymbol{g}_t} \xleftarrow{\boldsymbol{\mathcal{D}}} \boldsymbol{g}_t$
        \State $\boldsymbol{m}_t \gets \beta_1\boldsymbol{m}_{t-1}+\left( 1-\beta _1 \right) \boldsymbol{g}_{t}$
        \State $\boldsymbol{v}_t\gets \beta _2\boldsymbol{v}_{t-1}+\left( 1-\beta _2 \right) \boldsymbol{g}_{t}^{2}$
        \State $\hat{\boldsymbol{m}}_t\gets {\boldsymbol{m}_t}/{\left( 1-\beta _{1}^{t} \right)}$
        \State $\hat{\boldsymbol{v}}_t\gets {\boldsymbol{v}_t / {\left( 1-\beta _{2}^{t} \right)}}$
        \State $\Bar{\boldsymbol{\theta}}_t \gets \Bar{\boldsymbol{\theta}}_{t-1}-\alpha \cdot \hat{\boldsymbol{m}}_t/\sqrt{\hat{\boldsymbol{v}}_t+\boldsymbol{\epsilon }}$
        \State $\boldsymbol{\theta} \xleftarrow{\boldsymbol{\mathcal{D}}} \Bar{\boldsymbol{\theta}}$
    \EndWhile
    \State \textbf{return} $\boldsymbol{\theta}$
\end{algorithmic}
\end{algorithm}

\section{Results}

\subsection{Endogenous cognition recognizability in LLMs}

This study first addresses whether pre-trained LLMs already allocate task-relevant subnetworks.
We use the AM-DeepSeek-R1-Distilled-1.4M \cite{zhao202514millionopensourcedistilled} instruction dataset (distilled from DeepSeek-R1) to identify cognition-relevant subnetworks via a gradient attribution method as defined in Fig.\ref{fig:framework}b.
Specifically,
all instructions are annotated according to Bloom's taxonomy, as illustrated in Fig.\ref{fig:framework}a.
We then use these cognition-annotated instructions to identify isolated subnetworks in the LLMs.
The overall procedure is illustrated in Fig.\ref{fig:framework}b.

\begin{figure*}[htbp]
	\centering
	\includegraphics[width=1.0\textwidth]{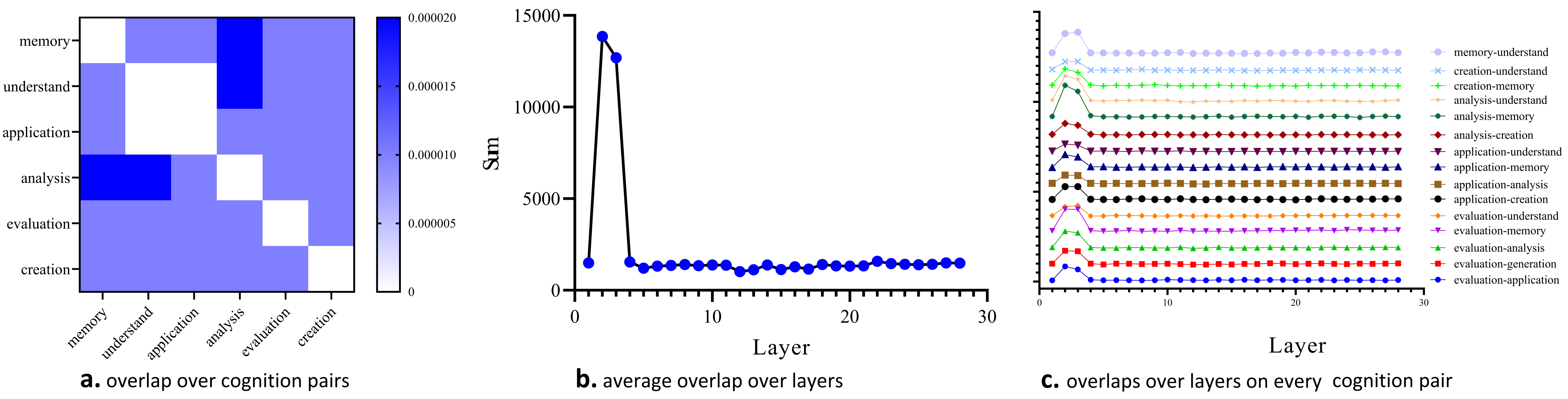}
	\caption{The cognition-isolated subnetwork overlap rate analysis. \textbf{a}, The average overlaps across all layers on all cognition pairs. \textbf{b}, The average overlap rates over all cognition pairs across all layers. \textbf{c}, The detailed overlaps over every cognition pair and all layers.}
	\label{fig:avgoverlap}
\end{figure*}

Fig.\ref{fig:avgoverlap}a reports the overlap rates between sparse subnetworks across all cognition pairs.
All overlaps are below $0.002\%$, suggesting that pre-trained LLMs exhibit an endogenous partition of cognition.
This observation supports our motivation to develop cognition-isolated, memory-efficient fine-tuning in this study.

We further analyze how subnetwork overlaps vary across layers.
Fig.\ref{fig:avgoverlap}b shows the overlaps across representation layers,
where most layers remain at a similarly low level, except for the 2nd and 3rd layers,
which exhibit higher overlap.
To examine each cognition pair in detail,
Fig.\ref{fig:avgoverlap}c plots the layer-wise overlap curves for all cognition pairs.
The trends are consistent with the average results in Fig.\ref{fig:avgoverlap}b,
with pronounced overlap peaks at the 2nd and 3rd layers across cognition pairs.

Taken together, these results align with recent findings \cite{jin2025exploring, zhao2024layer, liu2024fantastic} that the first few layers of LLMs tend to encode more general representations, exhibiting weaker cognitive discrimination than deeper layers.
As depth increases, representations gradually become more task- and cognition-specific.

\subsection{Continual learning performance}

After identifying the cognition-isolated subnetworks, we propose a subnetwork-local back-propagation algorithm, termed cognition-isolated memory-efficient optimization (CiMEO).
To enable memory-efficient optimization over sparse subnetworks within dense LLMs, we reformulate AdamW into a state-serialized variant, as illustrated in Fig.\ref{fig:framework}c.
In addition, we pre-train a cognitive classifier on the cognition-annotated AM-DeepSeek-R1-Distilled-1.4M instruction dataset.
This classifier is then frozen during continual learning and is used to route each incoming sample to its corresponding subnetwork for parameter updates.

We evaluate CiMEO on the AM-DeepSeek-R1-Distilled-1.4M instruction dataset.
Baselines include full-parameter fine-tuning with AdamW \cite{iclrLoshchilovH19}, as well as PEFT methods such as LoRA \cite{hu2022lora}, MiSS \cite{kang2024balancing}, Bone \cite{kang2024bone}, Shira \cite{bhardwaj2024sparse}, Dora \cite{liu2024dora}, Eva \cite{paischerone}, Lorafa \cite{zhang2023lora}, Olora \cite{buyukakyuz2024olora}, Pissa \cite{meng2024pissa}, Road \cite{liao2024in}, and MEFT methods of AdaLOMO \cite{lv2024adalomo}, Galore \cite{zhao2024galore}, AdaRankGrad \cite{refaeladarankgrad} and ZO-AdaMU \cite{jiang2024zo}.
All compared methods use the same training setup: batch size $B=64$, a constant learning rate $r=1e-5$, and model parameters.
For PEFT methods, the low-rank dimension is set to $r=16$, and when an $\alpha$ parameter is used,
we set $\alpha=32$ for the query, key, value, output, gate, up and down projection weights in the Transformer.

All models are evaluated on five representative and challenging benchmarks.
(\romannumeral1) \textbf{HumanEval-CN} \cite{chen2021evaluating}, which measures code reasoning and syntax fidelity for executable code generated from Chinese prompts.
(\romannumeral2) \textbf{GaokaoBench\_2010} \cite{Zhang2023EvaluatingTP}, which assesses complex, multi-disciplinary reasoning and problem-solving on the 2010 Chinese National College Entrance Examination (Gaokao).
(\romannumeral3) \textbf{CMMLU} \cite{li2024cmmlu}, which tests multitask language understanding across 67 subjects spanning humanities, social sciences, STEM, and medicine.
(\romannumeral4) \textbf{CEVAL} \cite{huang2023ceval},
which emphasizes expert-level question answering across civil service, legal, medical, and other professional exams.
(\romannumeral5) \textbf{GSM8K} \cite{cobbe2021training}, a grade school math dataset focusing on arithmetic and basic logical reasoning.

% Table generated by Excel2LaTeX from sheet 'Sheet2'
\begin{table*}[htbp]
  \centering
  {\footnotesize
    \begin{tabular}{cp{8em}cccccc}
    \toprule
    \rowcolor[rgb]{ .847,  .839,  .761} \multicolumn{2}{c}{\multirow{2}[2]{*}{Models}} & humaneval\_cn & gaokao & cmmlu & ceval & gsm8k &  \\
    \rowcolor[rgb]{ .847,  .839,  .761} \multicolumn{2}{c}{} & \textcolor[rgb]{ .275,  .471,  .525}{pass@1} & Acc (\%) & Acc (\%) & Acc (\%) & Acc (\%) & Avg \\
    \midrule
    \multicolumn{2}{c}{Qwen3-1.5B-Base} & 4.27  & 5.14  & 0.02  & 58.01  & 66.87  & 26.86  \\
    \midrule
    Full-Parameter & Adamw & 3.05  & \textbf{11.42 } & 0.00  & \textbf{67.40 } & 39.27  & 24.23  \\
    \midrule
    \multirow{10}[20]{*}{PEFT} & \cellcolor[rgb]{ .925,  .918,  .875}Lora & \cellcolor[rgb]{ .925,  .918,  .875}3.66  & \cellcolor[rgb]{ .925,  .918,  .875}6.99  & \cellcolor[rgb]{ .925,  .918,  .875}0.02  & \cellcolor[rgb]{ .925,  .918,  .875}60.44  & \cellcolor[rgb]{ .925,  .918,  .875}\textbf{68.08 } & \cellcolor[rgb]{ .925,  .918,  .875}27.84  \\
\cmidrule{2-8}          & Miss  & 8.54  & 11.15  & 0.00  & 66.73  & 66.79  & 30.64  \\
\cmidrule{2-8}          & \cellcolor[rgb]{ .925,  .918,  .875}\textcolor[rgb]{ .031,  .031,  .031}{Bone} & \cellcolor[rgb]{ .925,  .918,  .875}9.76  & \cellcolor[rgb]{ .925,  .918,  .875}11.29  & \cellcolor[rgb]{ .925,  .918,  .875}0.00  & \cellcolor[rgb]{ .925,  .918,  .875}66.62  & \cellcolor[rgb]{ .925,  .918,  .875}66.87  & \cellcolor[rgb]{ .925,  .918,  .875}30.91  \\
\cmidrule{2-8}          & \textcolor[rgb]{ .031,  .031,  .031}{Shira} & 1.22  & 9.26  & 29.61  & 63.13  & 68.01  & 34.25  \\
\cmidrule{2-8}          & \cellcolor[rgb]{ .925,  .918,  .875}\textcolor[rgb]{ .031,  .031,  .031}{Dora} & \cellcolor[rgb]{ .925,  .918,  .875}0.00  & \cellcolor[rgb]{ .925,  .918,  .875}6.81  & \cellcolor[rgb]{ .925,  .918,  .875}19.58  & \cellcolor[rgb]{ .925,  .918,  .875}63.08  & \cellcolor[rgb]{ .925,  .918,  .875}66.26  & \cellcolor[rgb]{ .925,  .918,  .875}31.14  \\
\cmidrule{2-8}          & \textcolor[rgb]{ .031,  .031,  .031}{Eva} & 0.00  & 9.90  & 0.05  & 52.19  & 19.71  & 16.37  \\
\cmidrule{2-8}          & \cellcolor[rgb]{ .925,  .918,  .875}\textcolor[rgb]{ .031,  .031,  .031}{Lorafa} & \cellcolor[rgb]{ .925,  .918,  .875}6.10  & \cellcolor[rgb]{ .925,  .918,  .875}9.62  & \cellcolor[rgb]{ .925,  .918,  .875}\textbf{48.40 } & \cellcolor[rgb]{ .925,  .918,  .875}60.88  & \cellcolor[rgb]{ .925,  .918,  .875}66.03  & \cellcolor[rgb]{ .925,  .918,  .875}38.20  \\
\cmidrule{2-8}          & \textcolor[rgb]{ .031,  .031,  .031}{Olora} & \textbf{29.27 } & 10.91  & 0.09  & 67.06  & 35.10  & 28.49  \\
\cmidrule{2-8}          & \cellcolor[rgb]{ .925,  .918,  .875}\textcolor[rgb]{ .031,  .031,  .031}{Pissa} & \cellcolor[rgb]{ .925,  .918,  .875}3.05  & \cellcolor[rgb]{ .925,  .918,  .875}\textbf{11.42 } & \cellcolor[rgb]{ .925,  .918,  .875}0.00  & \cellcolor[rgb]{ .925,  .918,  .875}\textbf{67.40 } & \cellcolor[rgb]{ .925,  .918,  .875}39.27  & \cellcolor[rgb]{ .925,  .918,  .875}24.23  \\
\cmidrule{2-8}          & \textcolor[rgb]{ .031,  .031,  .031}{Road} & 5.49  & 8.85  & 10.78  & 59.48  & 67.40  & 30.40  \\
    \midrule
    \multirow{4}[8]{*}{MEFT} & \cellcolor[rgb]{ .925,  .918,  .875}AdaLOMO & \cellcolor[rgb]{ .925,  .918,  .875}5.49  & \cellcolor[rgb]{ .925,  .918,  .875}7.58  & \cellcolor[rgb]{ .925,  .918,  .875}0.85  & \cellcolor[rgb]{ .925,  .918,  .875}58.87  & \cellcolor[rgb]{ .925,  .918,  .875}67.32  & \cellcolor[rgb]{ .925,  .918,  .875}28.02  \\
\cmidrule{2-8}          & Gelore & 13.41  & 9.00  & 16.44  & 63.00  & 66.72  & 33.71  \\
\cmidrule{2-8}          & \cellcolor[rgb]{ .925,  .918,  .875}AdaRankGrad & \cellcolor[rgb]{ .925,  .918,  .875}14.02  & \cellcolor[rgb]{ .925,  .918,  .875}5.51  & \cellcolor[rgb]{ .925,  .918,  .875}17.66  & \cellcolor[rgb]{ .925,  .918,  .875}65.24  & \cellcolor[rgb]{ .925,  .918,  .875}66.49  & \cellcolor[rgb]{ .925,  .918,  .875}33.78  \\
\cmidrule{2-8}          & ZO-AdaMU & 3.05  & 10.54  & 5.18  & 64.26  & 67.25  & 30.06  \\
    \midrule
    \multirow{3}[6]{*}{Ours} & \cellcolor[rgb]{ .925,  .918,  .875}CiMEO & \cellcolor[rgb]{ .925,  .918,  .875}25.00  & \cellcolor[rgb]{ .925,  .918,  .875}6.98  & \cellcolor[rgb]{ .925,  .918,  .875}28.85  & \cellcolor[rgb]{ .925,  .918,  .875}66.22  & \cellcolor[rgb]{ .925,  .918,  .875}67.55  & \cellcolor[rgb]{ .925,  .918,  .875}\textbf{38.92 } \\
\cmidrule{2-8}          & \quad w/ min\_grad & 6.71  & 5.28  & 0.01  & 58.17  & 66.64  & 27.36  \\
\cmidrule{2-8}          & \quad \cellcolor[rgb]{ .925,  .918,  .875}w/ random\_grad & \cellcolor[rgb]{ .925,  .918,  .875}0.61  & \cellcolor[rgb]{ .925,  .918,  .875}4.28  & \cellcolor[rgb]{ .925,  .918,  .875}0.00  & \cellcolor[rgb]{ .925,  .918,  .875}57.24  & \cellcolor[rgb]{ .925,  .918,  .875}67.63  & \cellcolor[rgb]{ .925,  .918,  .875}25.95  \\
    \bottomrule
    \end{tabular}}%
    \caption{The testing results on 5 benchmarks of fine-tuned Qwen3-1.5B with different continual learning methods. The metric for HumanEval-CN uses pass@1, and other four benchmarks adopt percentage of accurate predictions.}
  	\label{tab:seqcl}%
\end{table*}%

Table \ref{tab:seqcl} summarizes the results on the five benchmarks.
Although CiMEO does not achieve the top score on any single benchmark,
it attains the best overall balance across tasks,
yielding the highest average performance (38.92\%).
On humaneval\_cn benchmarks, Olora achieves the best score (29.27\%), followed by CiMEO (25.00\%).
On the GaoKaoBench\_2010, CiMEO reaches 6.98\%, comparable to LoRA.
On CMMLU, CiMEO improves by 28.85\% over the baseline.
On CEVAL and GSM8K, CiMEO performs nearly as well as the top-1 methods.
Overall,
CiMEO provides consistent gains across diverse evaluation settings,
rather than excelling on a single benchmark.

We further conduct an ablation study comparing minimum gradient attribution (CiMEO w/ min) and random sampling (CiMEO w/ random) strategies against the default maximum gradient attribution strategy.
As shown in Table \ref{tab:seqcl},
results on 4 benchmarks (HumanEval\-CN, GaoKaoBench\_2010, CMMLU, and C-Eval) indicate that CiMEO with maximum gradient attribution substantially outperforms both ablated variants.
The above results support the core assumption of CiMEO: gradient magnitudes reflect neuron sensitivity to different cognitive stimuli and can therefore guide effective subnetwork selection.

These results suggest that CiMEO improves general adaptability and long-horizon stability in continual learning, rather than overfitting to a specific dataset or domain.
Such balanced gains are especially important in continual learning, as they indicate better generalization and reduced catastrophic forgetting under sequential exposure to diverse skills.
The observed cross-task improvements further imply enhanced parameter reuse and representation sharing, which are critical when tasks arrive incrementally.

Overall, the experiments highlight CiMEO’s potential to improve continual learning by (\romannumeral1) promoting more uniform performance across heterogeneous tasks, (\romannumeral2) supporting knowledge accumulation and transfer across domains, and (\romannumeral3) mitigating performance degradation caused by task interference.

\subsection{Stability-plasticity trade-off performance} 

\begin{figure*}[htbp]
	\centering
	\includegraphics[width=1.0\textwidth]{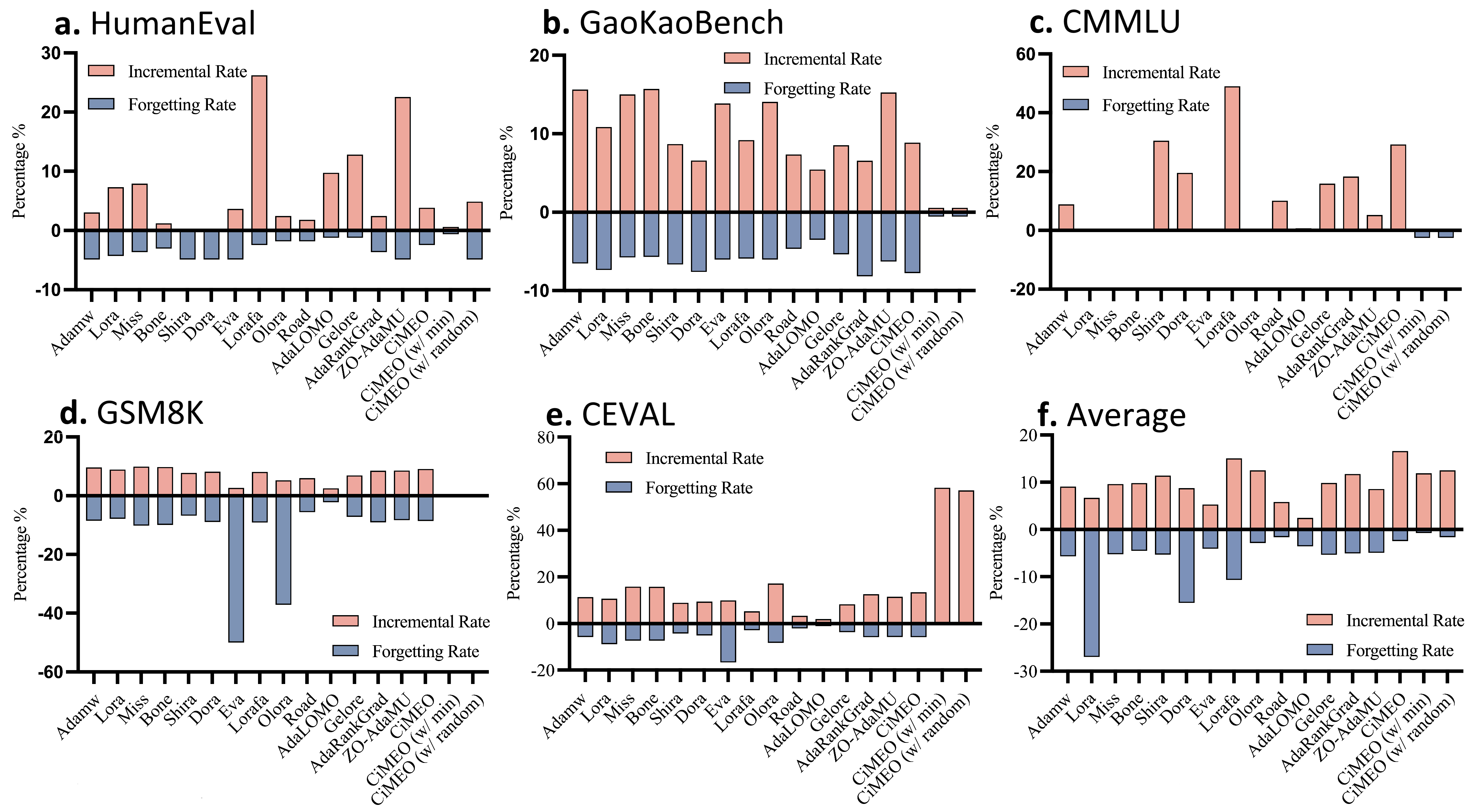}
	\caption{The forgetting and incremental rates on five benchmarks for all compared baselines. (\textbf{a}) for HumanEval-CN, (\textbf{b}) for GaoKaoBench-2010, (\textbf{c}) for CMMLU, (\textbf{d}) for GSM8K, (\textbf{e}) for CEval, and (\textbf{f}) is the average over all five benchmarks.}
	\label{fig:forgetting}
\end{figure*}

Anti-forgetting and incremental learning are two key metrics for evaluating continual learning.
In this study, we quantify them on specific benchmarks as follows.

\begin{equation}
	\begin{aligned}
		\mathcal{r}_{\textit{f}} &= \frac{1}{N} \sum_{n=1}^N \mathbb{I} \left[ \left( \hat{y}^{(n)}_{\textit{chkpt}} \neq \left( y^{(n)} \right) \& \hat{y}^{(n)}_{\textit{base}} = y^{(n)} \right) \right] \\
		\mathcal{r}_{\textit{i}} &= \frac{1}{N} \sum_{n=1}^N \mathbb{I} \left[ \left( \hat{y}^{(n)}_{\textit{chkpt}} = y^{(n)} \right) \& \left( \hat{y}^{(n)}_{\textit{base}} \neq y^{(n)} \right) \right]
	\end{aligned}
\end{equation}
where $\mathcal{r}_{\textit{f}}$ denotes the forgetting rate,
and $\mathcal{r}_{\textit{i}}$ denotes the incremental learning rate.
Here, $\hat{y}_{\textit{chkpt}}$ represents the predictions of the fine-tuned LLMs, and $\hat{y}_{\textit{base}}$ denotes the predictions of the base model.

Results are summarized in Fig.\ref{fig:forgetting} as bar charts.
Across HumanEval-CN, CMMLU, and GSM8K (Fig.\ref{fig:forgetting}a, Fig.\ref{fig:forgetting}c and Fig.\ref{fig:forgetting}d),
CiMEO exhibits consistently low forgetting rates, and its average forgetting rate (Fig.\ref{fig:forgetting}f) is close to the best-performing methods.
Meanwhile, CiMEO achieves strong incremental learning performance across benchmarks, ranking among the top methods overall.
Notably, although AdaLOMO and AdaRankGrad show relatively low forgetting, their incremental gains are limited, indicating insufficient plasticity.
In contrast, CiMEO provides a more favorable trade-off between stability (anti-forgetting) and plasticity (incremental learning).

We further observe that the ablated variants, CiMEO w/ min and CiMEO w/ random, can achieve very low forgetting on some benchmarks (e.g., GaokaoBench, GSM8K, and C-Eval), but they generally suffer from low incremental learning rates across tasks.
This suggests that subnetworks selected by minimal gradient attribution or random sampling do not effectively capture cognition-relevant parameters for continual adaptation.
Conversely, these findings support our core hypothesis that LLMs exhibit endogenous cognition isolation in their internal networks, and gradient magnitudes provide a practical signal for identifying and leveraging such isolated subnetworks.

\subsection{Weight stability analysis}

In addition to benchmark results, we analyze how different optimization methods affect representational transformations by examining the weight space through singular value decomposition (SVD).
The notation is defined as follows,

\begin{itemize}
	\item Continual learning checkpoints are indexed by $t=0,1,\dots,T$, where $t=0$ denotes the base model.
	\item We analyze all 2D weight matrices in each layer $1 \leq l \leq L$.
	\item Taking the query projection in self-attention as an example, the weight matrix at layer $l$ and checkpoint $t$ is denoted by $\mathbf{Q}_{l}^{(t)}$. Its SVD is $\mathbf{Q}^{(t)_l}=\mathbf{U}^{(t)}_{q,l} \boldsymbol{\Sigma}^{(t)}_{q,l} \mathbf{V}^{(t)}_{q,l}$, where $\boldsymbol{\Sigma}^{(t)}_{q,l}=\operatorname{diag}(\sigma_{1}\ge \sigma_{2}\ge \cdots \ge 0)$. The top-k left singular vectors are $\mathbf{U}^{(t)}_{q,l}[:,1\!:\!k] \in \mathbb{R}^{d_{\text{out}}\times k}$, and top-k right singular vector is $\mathbf{V}^{(t)}_{q,l}\left[ 1:k, : \right] \in \mathbb{R}^{k \times d_{\text{out}}}$.
\end{itemize}

\textbf{Subspace Overlap Analysis:}
For each linear projection matrix (e.g., the query projection) in each layer,
the top-k left singular vectors span a k-dimensional subspace capturing the dominant directions of the transformation.
We quantify the similarity between the base model $\mathbf{W}_0$ and a fine-tuned checkpoint $\mathbf{W}_t$ using the subspace overlap metric $\mathcal{O}(\mathbf{W}_0, \mathbf{W}_t; k)$,
\begin{equation}
	\mathcal{O}(\mathbf{W}_0, \mathbf{W}_t; k) = \frac{1}{k}\| \mathbf{Q}_0^\top \mathbf{Q}_t\|_F^2
\end{equation}
where $\mathbf{Q}_0$ and $\mathbf{Q}_t$ denote the top-k orthonormal singular-vector matrices of $\mathbf{W}_0$ and $\mathbf{W}_t$, respectively,
and $\|\cdot\|_F$ denotes the Frobenius norm.
By definition,
$\mathcal{O}(\mathbf{W}_0, \mathbf{W}_t; k) \in \lbrack 0,1 \rbrack$, where $1$ indicates identical subspaces and $0$ indicates orthogonal subspaces.

\begin{figure*}[htbp]
	\centering
	\includegraphics[width=1.0\textwidth]{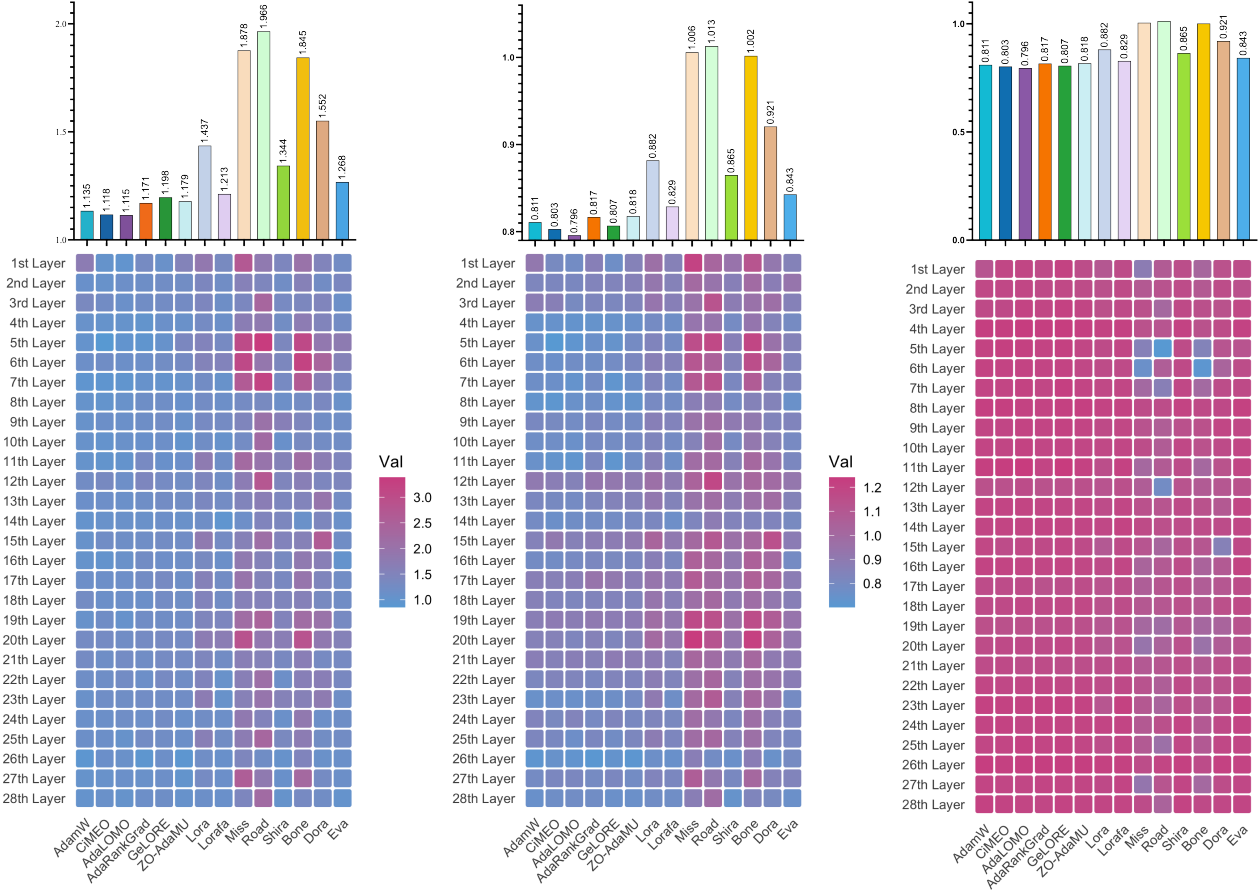}
	\caption{The weight stability analysis. (\textbf{a}), The maximum principle angles between Qwen3-1.5B and all compared methods fine-tuned checkpoints across all layers, and (\textbf{d}) average over layers. (\textbf{b}), The maximum principle angles between Qwen3-1.5B and all compared methods fine-tuned checkpoints across all layers, and (\textbf{e}) average over layers. (\textbf{c}), the subspace overlaps between Qwen3-1.5B and different methods fine-tuned checkpoints across all layers, and (\textbf{f}) average over layers.}
	\label{fig:singular}
\end{figure*}

We compute the average overlap across all layers and checkpoints, and summarize the results in Fig.\ref{fig:singular}.
A higher overlap indicates stronger subspace stability and more minor changes in the learned representations.
Overall, CiMEO achieves high overlap values comparable to the most stable baselines, suggesting that it largely preserves the dominant weight subspaces during continual learning.
This observation is consistent with the anti-forgetting results in Fig.\ref{fig:forgetting},
where methods with higher subspace overlap (e.g., CiMEO) tend to exhibit stronger resistance to forgetting.

\textbf{Principal-angle analysis of weight spaces:} Principal angles characterize the rotation of the dominant subspaces between the base model and fine-tuned checkpoints.
For each layer $l$ at checkpoint $t$,
we define the principal subspaces using the top-$k$ left singular vectors $\mathbf{Q}_l^{(0)} = \mathbf{U}^{(0)}_l[:, 1\!:\!k]$ and $\mathbf{Q}^{(t)}_l = \mathbf{U}^{(t)}_l[:, 1\!:\!k]$.
We compute the cosine similarity $\boldsymbol{\sigma}^{(t)}_l = \left( \mathbf{Q}^{(0)}_l \right)^T \left( \mathbf{Q}^{(t)}_l \right)$,
and define the $i$-th principal angle as $\theta_{l,i}^{(t)}=\arccos \left( \sigma_{l,i}^{(t)} \right)$, for $i=1,\cdots,k$.
The mean and maximum principal angles for the $l$-th layer as follows,
\begin{equation}
	\hat{\theta}^{(t)}_l = \frac{1}{k} \sum_{i=1}^{k} \theta^{(t)}_{l,i} \quad
	\acute{\theta}^{(t)}_l = \max_i \lbrace \theta^{(t)}_{l,1}, \cdots, \lbrace \theta^{(t)}_{l,k} \rbrace
\end{equation}

Fig.\ref{fig:singular}a and \ref{fig:singular}b report the layer-wise principal angles and the corresponding mean/max summaries across layers.
These results reveal which layers maintain stable transformations and which undergo substantial rotations during continual learning.
CiMEO exhibits relatively small mean and maximum principal angles, indicating that it induces only minor subspace rotations.
The above results suggest that CiMEO primarily builds on the original feature structure rather than substantially reorienting it, providing a mechanistic explanation for its strong anti-forgetting performance.

\section{Conclusion}
\label{sec:conclusion}

In this work, we investigated functional subnetwork localization for mitigating catastrophic forgetting in continual learning of large language models.
Gradient-based attribution revealed distinct parameter subsets associated with six cognitive categories derived from Bloom's taxonomy.
Building on this observation, we proposed CiMEO, which integrates cognition-guided routing, attribution-based parameter selection, and state-serialized AdamW optimization to enable localized adaptation while reducing optimizer-state requirements.

Experiments across 5 benchmarks demonstrated that CiMEO achieves the highest average performance among the compared methods, balancing knowledge acquisition and retention.
Ablation studies validated the importance of subnetwork selection, while weight-space analyses indicated substantial preservation of pre-trained representations.
These findings highlight functional subnetwork localization as a promising approach to efficient and stable continual learning, suggesting that cognition-guided parameter updates can mitigate interference without extensive modifications to pre-trained models.

\section{Acknowledgments}
\noindent This work is jointly supported by grants from the Shenzhen Science and Technology Program (No.KCXFZ20230731093001002, No.KJZD20230923115113026), Shenzhen Basic Research General Program (No.JCYJ20250604145438051), Guangdong Basic and Applied Basic Research Foundation (No.2026A1515012168, No.2023A1515110561, No.2025A1515012263), Establishment Project of Shenzhen Chinese Calligraphy Digital Simulation Technology Engineering Laboratory (No.XMHT20190108009), Shenzhen Key Laboratory of Intelligent Computing in Network Environment, the National Natural Science Foundation of China (No.62276075, No.62506099), the Postdoctoral Fellowship Program of CPSF (No.GZC20252738, No.GZC20252759).

% IEEE Bibliography
\bibliographystyle{IEEEtran}
\bibliography{sn-bibliography}

\iffalse
\begin{IEEEbiographynophoto}{Jane Doe}
Biography text here without a photo.
\end{IEEEbiographynophoto}

\begin{IEEEbiography}[{\includegraphics[width=1in,height=1.25in,clip,keepaspectratio]{fig1.png}}]{IEEE Publications Technology Team}
In this paragraph you can place your educational, professional background and research and other interests.\end{IEEEbiography}
\fi

\end{document}